# ENHANCED INPUT MODELING FOR CONSTRUCTION SIMULATION USING BAYESIAN DEEP NEURAL NETWORKS

Yitong Li
Wenying Ji

Department of Civil, Environmental, and Infrastructure Engineering
George Mason University
Fairfax, VA 22030, USA

## ABSTRACT

This paper aims to propose a novel deep learning-integrated framework for deriving reliable simulation input models through incorporating multi-source information. The framework sources and extracts multi-source data generated from construction operations, which provides rich information for input modeling. The framework implements Bayesian deep neural networks to facilitate the purpose of incorporating richer information in input modeling. A case study on road paving operation is performed to test the feasibility and applicability of the proposed framework. Overall, this research enhances input modeling by deriving detailed input models, thereby, augmenting the decision-making processes in construction operations. This research also sheds lights on prompting data-driven simulation through incorporating machine learning techniques.

## 1 INTRODUCTION

Construction operations are widely known for its uncertain characteristics, such as varying weather conditions, equipment breakdowns, and management interference (AbouRizk et al. 1994). Due to the fact that these uncertainties severely impact the decision-making processes, it is essential to model construction operations as random processes, which are typically represented by random variables sampled from a fitted probabilistic distribution using real-world data samples. While the fitted probabilistic distribution is capable of representing uncertainties involved in construction operations, its representativeness may not be reliable due to the exclusion of detailed information in input modeling.

To better illustrate this limitation, an example on modeling truck hauling duration is provided in Figure 1. Figure 1(a) demonstrates data samples that were collected for the truck hauling operation under three weather conditions (i.e., rainy, windy, and sunny). In Figure 1(b), the outer input model (in black color) represents the truck hauling duration without considering weather conditions; while the inner ones represent the truck hauling duration under each weather condition. Apparently, the inner three input models are more representative to certain weather conditions (i.e., rainy, windy, and sunny) compared to the outer one, which eliminates the occasion of sampling an erroneous random variable from undesired weather conditions—sampled random variables (i.e., green dots) for the windy weather condition are erroneously sampled from rainy and sunny weather conditions. Therefore, to reduce the inaccuracy associated with the random variables sampled from undesired conditions (i.e., input models), extra information needs to be incorporated to derive reliable input models at detailed levels.

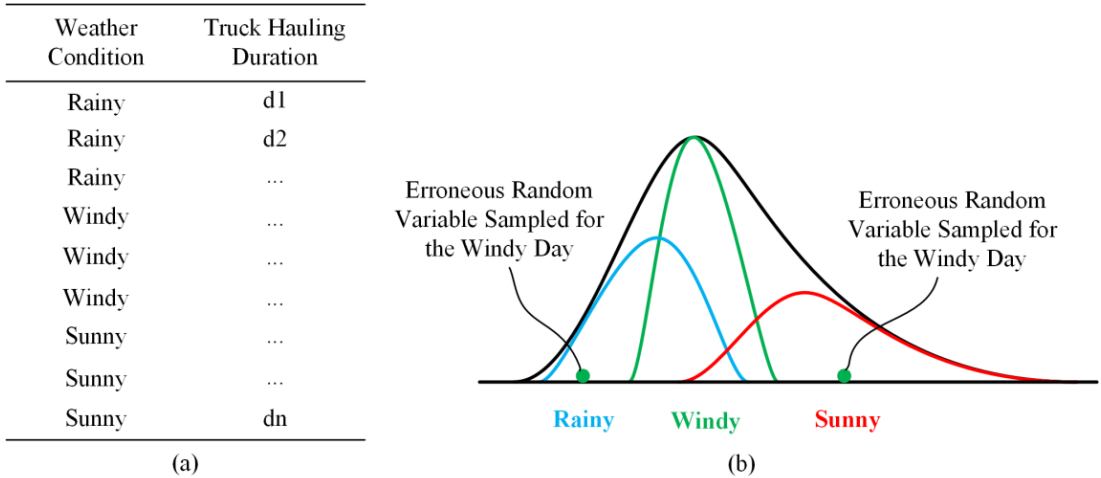

Figure 1: Illustrative example of truck hauling duration input modeling.

In the above example, the weather is solely considered for the illustration purpose. In actual construction operations, various types of information are becoming available due to the implementation of advanced technologies (e.g., sensors and information systems) and these types of information have the potential to be incorporated for deriving reliable input models. Although promising, derivation of detailed simulation input models using these types of information has yet been addressed mainly due to the lack of (1) a capable fusion strategy to transform this information into useable data for advanced analytical analysis and (2) a quantitative approach to utilize such data for enhanced input modeling.

To address these issues, this research aims to propose a Bayesian deep neural network-based framework for enhanced simulation input modeling. Specifically, this objective is achieved through (1) adapting multi-source information to enrich the dataset used for input modeling and (2) implementing Bayesian deep neural networks to derive detailed input models using the adapted information. The remainder of this paper is arranged as follows. In the next section, motivations related to applying Bayesian deep neural networks in input modeling are discussed. After that, the main components of the proposed framework including data source, data adapter, and Bayesian deep neural network are introduced. Following the framework, a case study on road paving operation is provided to demonstrate the feasibility and applicability of the proposed framework. In the end, contributions, limitations, and future work are concluded.

## 2    RATIONALE

Previously, multiple studies have been conducted on developing Bayesian-based input modeling approaches for various construction applications. Notable examples include improvement of construction fabrication quality control system through a Bayesian inference-based quality performance modeling (Ji and AbouRizk 2018), estimation of quality-induced rework cost by implementing the previously developed quality performance modeling approach (Ji and AbouRizk 2018), and prediction of ground conditions during tunneling operations via a Bayesian-based Hidden Markov Model (Werner et al. 2018). Although, an attempt has been further conducted to generalize the Bayesian-based input modeling strategy using Markov Chain Monte Carlo (MCMC)-based numerical approximation (Wu et al. 2019), the capabilities of these models are still limited to univariable input modeling. In other words, these methods only used one type of information (i.e., single input) to derive input models, which are insufficient for obtaining reliable input models. Therefore, an approach, which is capable of incorporating multiple types of information in input modeling as well as representing uncertainties associated with construction operations, is required for deriving input models at detailed levels.

Deep learning, which is recognized as a new domain of machine learning, is well-known for its ability to discover intricate structures in high-dimensional data (LeCun et al. 2015). Although capable of integrating multiple inputs for deriving the output, the output obtained from deep learning methods is incapable of representing the desired uncertainties for the input model since the output is given as a deterministic value (i.e., point estimate). To address this limitation, a novel Bayesian inference-based deep neural network, which is capable of integrating multiple inputs for deriving the output as well as modeling uncertainties of the estimated output, has been developed and validated (Kendall and Gal 2017; Gal and Ghahramani 2016). Compared to deep learning methods, Bayesian deep neural networks are capable of modeling two types of uncertainties (i.e., epistemic uncertainty and aleatoric uncertainty). In detail, epistemic uncertainty is a type of uncertainty arises due to the lack of knowledge, it can be explained away by gathering more information, while aleatoric uncertainty represents intrinsic randomness of a phenomenon that cannot be reduced even if more information were to be collected (Der Kiureghian and Ditlevsen 2009). In the context of input modeling for construction simulation, aleatoric uncertainty is particularly crucial as it well represents uncertain situations that happen during construction operations. Therefore, in this research, only aleatoric uncertainty is included in deriving input models through the Bayesian deep neural network approach. Detailed explanations and mathematical derivations on the Bayesian deep neural network are referred to the publications (Kendall and Gal 2017; Gal and Ghahramani 2016).

## 3 METHODOLOGY

To achieve the goal of deriving reliable input models by incorporating multi-source information using Bayesian deep neural networks, a framework containing four components (i.e., data source, data adapter, Bayesian deep neural network, and input modeling) is proposed, shown in Figure 2. First, raw data sources related to construction operations are identified to enrich the information that is used for input modeling. Then, a data adapter is used to transform raw data sources into an interpretable and compatible dataset through processes of data connection, data mapping, and data cleaning. After that, a Bayesian deep neural network, which takes the transformed dataset as input to derive a probabilistic distribution for the desired output (i.e., input model), is applied to prepare input models for simulation purposes.

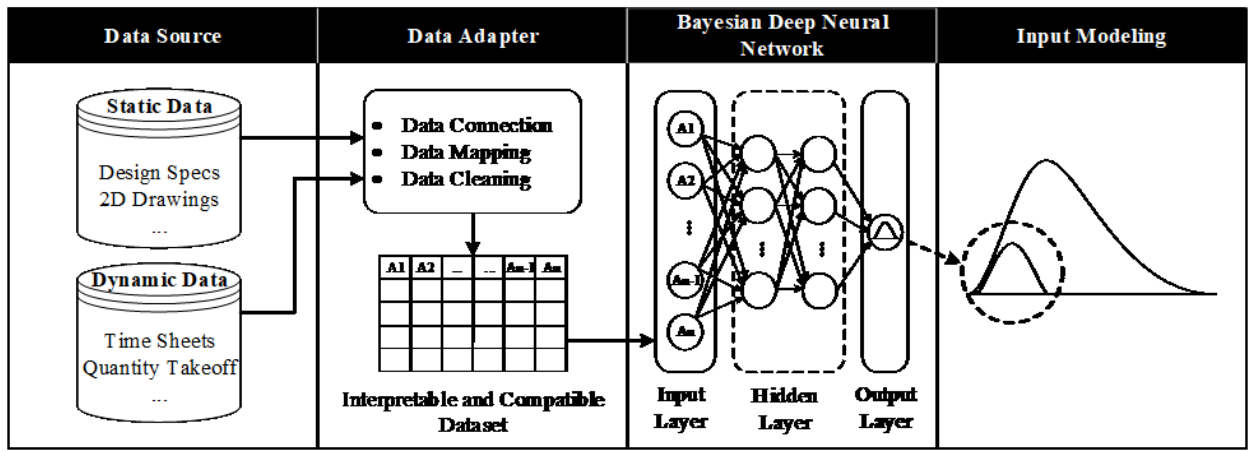

Figure 2: Research framework.

### 3.1 Data Source

The construction industry has been dealing with vast amounts of data generated from various data sources (Bilal et al. 2016). The multi-source data comprises potential values that can be extracted to enhance decision-making processes in the construction industry (Hovnanian et al. 2019). Common types of data in the construction industry are categorized into two groups, namely, dynamic data and static data (Ji and AbouRizk 2018). Dynamic data is archival, periodically updated data that is commonly stored in the entrepreneurial resource planning systems and standalone applications (e.g., timesheets, material takeoff quantities, and safety incidents). Static data is the pre-defined data which is commonly stored in engineering design systems such as design specifications, two-dimensional (2D), and three-dimensional (3D) drawings (Ji and AbouRizk 2018). However, these data sources are typically isolated and cannot be directly utilized for decision support. To address this issue, these isolated data sources need to be firstly mapped and then transformed into a clean and tidy format prior to further analysis.

### 3.2 Data Adapter

Although a vast amount of information has been collected, construction companies are still struggling with extracting useful insights from the information due to the challenge of transforming the sparsely stored information into an aggregated and interpretable table (Hovnanian et al. 2019). As such, a data adapter, which aggregates and transforms information from various data sources into one compatible and interpretable dataset, is required. Here, a data adapter mainly composes three functionalities: (1) data connection, which functions to connect data from various sources and formats, (2) data wrangling, which reshapes, groups, and combines data from multiple sources, and (3) data cleaning, which handles missing values and outliers (Ji and AbouRizk 2018). Once multi-source data has been transformed into useable information at the required level, intelligent algorithms (i.e., Bayesian deep neural networks) can be used to generate reliable input models for simulation.

### 3.3 Bayesian Deep Neural Network

To interpret the non-linear relationship between inputs (i.e., multi-source data) and outputs (i.e., input models), the Bayesian deep neural network (Kendall and Gal 2017) is utilized. This approach delivers a distribution for the desired output (i.e., input model) to include both the input-dependent aleatoric uncertainty and epistemic uncertainty. In this well-established approach, aleatoric uncertainty is suitable to reflect uncertain conditions of construction operations due to its representativeness of the random variability among the samples of a population. Therefore, here, only aleatoric uncertainty is included in the output derived from the Bayesian deep neural network to prepare the input model. In the developed approach, aleatoric uncertainty is modeled by placing a distribution over the output of the model and expressed using mean and variance (Kendall and Gal 2017). The analytical derivation is referred to the publication (Kendall and Gal 2017).

### 3.4 Input Modeling

In the context of construction simulation, input modeling using Bayesian deep neural networks provides means to incorporate multi-source data for deriving detailed input models through the following steps: (1) training a model that interprets the association between inputs (i.e., adapted dataset) and outputs (i.e., input models) using a Bayesian deep neural network and (2) deriving the desired output distribution given a certain set of inputs using the trained model. The derived output distribution is then utilized as an input model for simulation.

## 4 CASE STUDY

In this section, a case study on a road paving operation is conducted to test the feasibility and applicability of the proposed framework. In the road paving operation, concrete is first hauled from a batch plant to a slip form paving machine using dump trucks. Then, the paving machine forms and finishes one lane of plain concrete pavement as it continuously moves forward. In some cases, a spreader is used in front of the paving machine to assist the paving process. During the road paving operation, road paving operation performance (i.e., completion time) is significantly affected by the operation condition. Therefore, deriving a reliable input model, which quantifies the paving operation productivity based on the road paving condition, is required. In the following sections, the detailed processes of data preparation and input modeling using Bayesian deep neural networks are described.

## 4.1   Data Preparation

In the presented study, the general data preparation process follows procedures described in the methodology section. Here, detailed data source identification and data adaption are not discussed since data collection and storage methods vary among projects. To demonstrate the prepared data, a sample dataset is shown as Table 1. This dataset shows the paving operation productivity (column 1) and multiple variables (columns 2-10) that define the operation condition. These variables include: (1) weather—air temperature and air humidity; (2) site condition—traffic congestion (Yes/No), use of spreader (Yes/No), and age of the paving machine (Years from 1 to 5); (3) survey information—roadway slope and curvature; and (4) concrete design information—concrete slump and air entrainment. The final prepared dataset contains 406 rows, in which, each row represents one road paving scenario under a certain operation condition represented by the 9 attributes shown in columns 2-10. In the following step, the prepared dataset will serve as the input for training and testing the Bayesian deep neural network model.

Table 1: Sample dataset of the road paving activity.

| Productivity | Slump | Congestion | Spreader | Air Entrainment | Temperature | Humidity | Slope | Curvature | PaverAge |
|---|---|---|---|---|---|---|---|---|---|
| $m^3/hr$ | cm | Yes(1)/No(0) | Yes(1)/No(0) | % | °C | % | % | 1/m | Year |
| 66.08 | 3.0 | 1 | 0 | 4.6 | 5.3 | 59.6 | -0.7880 | 0.001 | 4.0 |
| 86.79 | 4.6 | 0 | 0 | 4.2 | 27.2 | 82.4 | 0.4117 | 0.001 | 2.5 |
| 69.35 | 4.0 | 1 | 0 | 4.0 | 9.3 | 70.1 | 1.4199 | 0.000 | 2.5 |
| 93.69 | 4.5 | 0 | 0 | 4.3 | 23.6 | 73.1 | 0.8632 | 0.001 | 1.0 |
| ... | ... | ... | ... | ... | ... | ... | ... | ... | ... |

## 4.2   Input Modeling Using Bayesian Deep Neural Network

Here, input modeling using Bayesian deep neural networks was implemented in R (R Core Team 2018) using packages Keras (Allaire and Chollet 2018) and TensorFlow (Allaire and Tang 2018) by following the above-mentioned two steps in Section 3.4. During the training process, 80% of the adapted dataset was used as the training set. Three hidden layers together with the rectified linear unit (ReLU) activation function were set for the Bayesian deep neural network model. The network weights (i.e., associations between inputs and outputs) were analyzed via the standard back-propagation procedure following the defined loss function (Kendall and Gal 2017) and the Adaptive Moment Estimation (Adam) optimization function.

After the training process, the trained model was applied to the remaining 20% test set to validate the model accuracy. In the validation process, imputing each set of 9 attributes to the model leads to a probabilistic distribution that represents uncertainties of paving operation productivity. The validation result is illustrated in Figure 3, where the x-axis represents the observed productivity and the y-axis represents the predicted productivity. The grey vertical bar represents a 95% confidence interval ($\mu \pm 1.96\sigma$; $\mu$ represents the mean and $\sigma$ represents the standard deviation) of the predicted productivity distribution. The diagonal line represents the situation when the predicted productivity distribution is able to cover the observed productivity. From Figure 3, it is observed that all the predicted 95% confidence intervals are intersected with the diagonal line, which validates the predicted productivity distribution using the Bayesian deep neural network is capable of representing the observed productivity. Furthermore, it is also observed that areas with dense data samples have a smaller confidence interval than areas with sparse data samples, which is in line with the fundamental statistical theory—uncertainty decreases as the sample size increases, and vice versa.

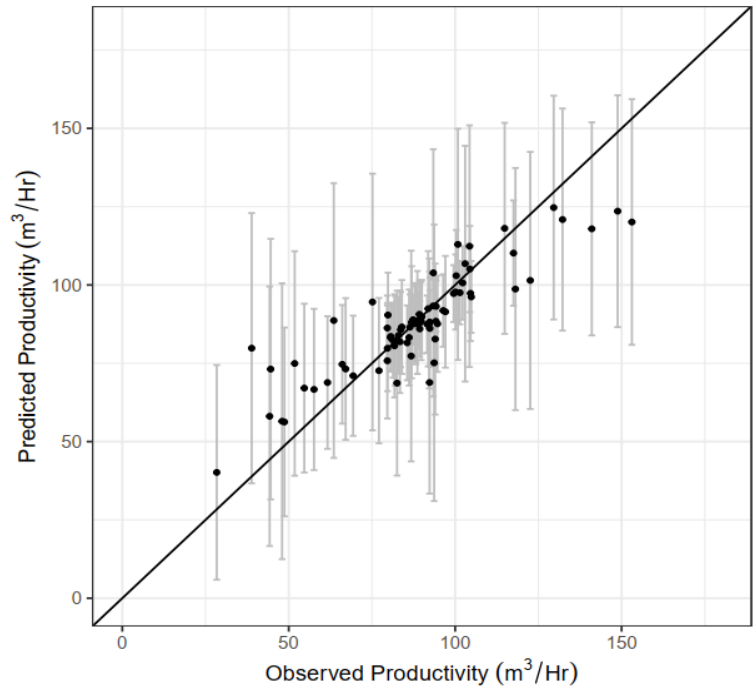

Figure 3: Comparison between observed productivity and predicted productivity.

To illustrate the final representation of the input model, three road paving scenarios with the best, medium, and worst productivities are selected from the dataset. Detailed operation scenarios are demonstrated in Table 2. Using one operation scenario (i.e., 9 variables) as the model's inputs, the trained model outputs an estimation of the paving operation productivity in the format of a probabilistic distribution (expressed using mean and variance). Similarly, applying the trained model to all three scenarios, three input models are obtained and shown in Figure 4.

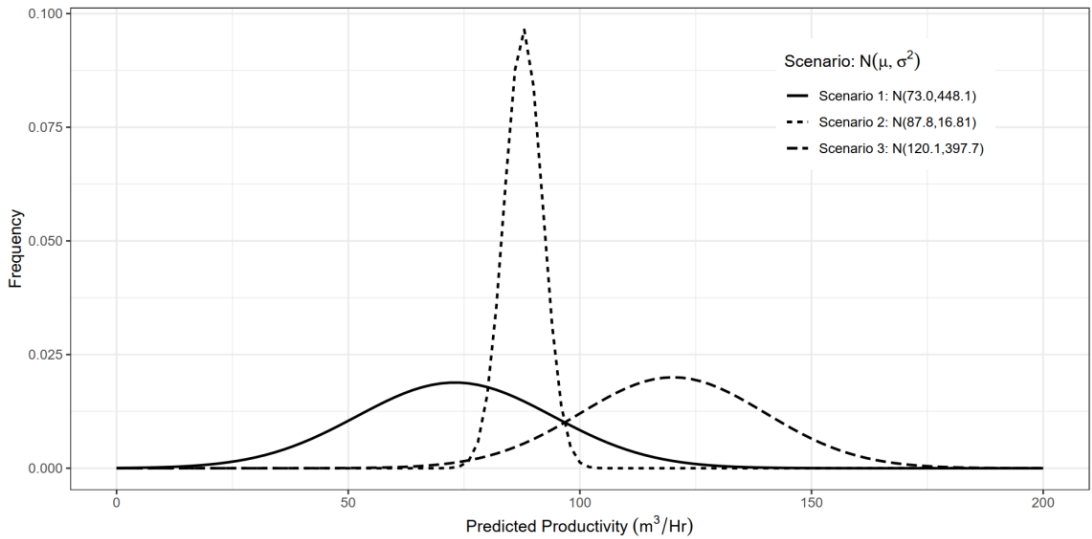

Figure 4: Input models of paving operation productivity under three scenarios.

Table 2: Working conditions of the road paving operation under three scenarios.

| Senarios | Slump cm | Congestion Yes(1)/No(0) | Spreader Yes(1)/No(0) | Air Entrainment % | Temperature °C | Humidity % | Slope % | Curvature 1/m | PaverAge Year |
|---|---|---|---|---|---|---|---|---|---|
| S1 | 4.5 | 1 | 0 | 4.5 | 6.5 | 84.6 | 0.0000 | 0.001 | 5.0 |
| S2 | 4.3 | 1 | 0 | 4.2 | 21.8 | 86.0 | -3.4952 | 0.001 | 2.5 |
| S3 | 3 | 0 | 1 | 4.5 | 7.7 | 60.1 | 1.2028 | -0.001 | 0.0 |

Refer to Table 2, the paving operation in scenario 3 has the highest productivity since a brand-new machine was used with the assistance of a spreader on an uncongested road. Compared to scenario 3, the productivity of the paving operation in scenario 2 has relatively lower productivity due to its 2.5-year usage and paving without spreader assistance on a congested road. Unsurprisingly, compared to scenarios 2 and 3, the paving operation in scenario 1 has the lowest productivity due to its 5-year usage and paving without spreader assistance on a congested road. Although the association between productivity and the three operation conditions is easy to interpret, a quantitative representation of the extent to which these attributes affect productivity is still unknown. Moreover, when more types of operation variables (e.g., slump, humidity, and slope) are involved, the association between these variables and productivity will become more complex and no longer be interpretable.

In summary, the Bayesian deep neural network, which is capable of modeling complex associations between multi-source information and input models, functions well to serve the input modeling purposes.

## 5   CONCLUSION

In this research, a Bayesian deep neural network-based framework was inventively proposed to enhance input modeling by detailing input models through incorporating multi-source information. The feasibility and applicability of the proposed framework were demonstrated through a road paving case study. While this framework is proposed to enhance simulation of construction operations, it is also applicable to other simulation applications that could benefit from using multi-source information. Although being capable of incorporating uncertainties, the proposed framework is limited to the derivation of normal distributions. Further research will be focused on extending the proposed framework to derive more flexible distributions for input modeling purposes. Further studies will also be conducted to measure the impact of the proposed framework on simulation performance.

## AUTHOR BIOGRAPHIES

**YITONG LI** is a Ph.D. student in the Department of Civil, Environmental & Infrastructure Engineering, George Mason University. Yitong's current research area fosuces on construction simulation input modeling. Her e-mail address is yli63@gmu.edu.

**WENYING JI** is an assistant professor in the Department of Civil, Environmental & Infrastructure Engineering, George Mason University. Dr. Ji received his PhD in Construction Engineering and Management from the University of Alberta. Dr. Ji is an interdisciplinary scholar focused on the integration of advanced data analytics, complex system simulation, and construction management to enhance the overall performance of infrastructure systems. His e-mail address is wji2@gmu.edu.